\documentclass[11pt]{article}
\PassOptionsToPackage{pdfusetitle}{hyperref}
\usepackage[preprint]{acl}
\usepackage{times}
\usepackage{latexsym}
\usepackage[T1]{fontenc}
\usepackage[utf8]{inputenc}
\usepackage{microtype}
\usepackage{inconsolata}
\usepackage{graphicx}
\usepackage{booktabs}
\usepackage{amsmath}
\usepackage{amssymb}
\usepackage{xcolor}
\usepackage{multirow}
\usepackage{url}


\newcommand{\ExternalNPerClass}{500}
\newcommand{\TrainNPerClass}{566}
\newcommand{\DevNPerClass}{100}

\newcommand{\BestMethodName}{Mean + KD}

\newcommand{\BaselineSuppressK}{4}
\newcommand{\BestSuppressK}{11}
\newcommand{\BaselineMMLU}{45.8\%}

\newcommand{\BaselineHellaSwag}{53.5\%}
\newcommand{\BestMMLU}{45.7\%}

\newcommand{\BestHellaSwag}{56.2\%}
\newcommand{\BaselineAdaptiveRankFourHarm}{31.2\%}

\newcommand{\BestAdaptiveRankFourHarm}{22.2\%}

\newcommand{\AllLayerBestSuppressK}{3}
\newcommand{\QwenBaselineSuppressK}{2}
\newcommand{\QwenBestSuppressK}{1}

\providecommand{\QwenBaselineSuppressK}{\textsc{pending}}
\providecommand{\QwenBestSuppressK}{\textsc{pending}}

\newcommand{\harm}{\mathrm{harm}}
\newcommand{\safe}{\mathrm{safe}}
\newcommand{\placeholderfigure}[1]{%
  \fbox{\parbox[c][3.1cm][c]{0.96\linewidth}{\centering #1}}%
}

\title{Does Representation Matching Make Refusal Harder to Ablate? A~Controlled Study of Iterated Linear Attacks}

\newcommand{\equalcontribution}{\texorpdfstring{\hyperlink{equal-contribution}{\textsuperscript{*}}}{}}
\author{Jonathan Merchan\equalcontribution, Deepanshu Mody\equalcontribution, and Acey Vogelstein\equalcontribution
  \texorpdfstring{\\ {\normalfont New York University}}{}}

\begin{document}
\maketitle
\begingroup
\renewcommand{\thefootnote}{\fnsymbol{footnote}}
\footnotetext[1]{\hypertarget{equal-contribution}{All authors contributed equally.}}
\endgroup

\begin{abstract}
Projecting a direction from the residual stream can suppress refusal in instruction-tuned language models.
We ask whether representation matching during fine-tuning delays suppression while keeping refusal on harmless prompts low.
We vary mean matching, covariance matching, distillation, and token pooling, and compare with Refusal Feature Adversarial Training (ReFAT).
Under an iterated protocol excluding the final 20\% of source layers, the first observed rank at which the pointwise 95\% Wilson upper bound for harmful-prompt refusal fell below 30\% was $K=4$ for the untouched Llama checkpoint and $K=11$ for mean matching with knowledge distillation (KD), selected on development data.
These crossings describe one checkpoint per condition.
Clean MMLU, GSM8K, and HellaSwag scores were similar or higher for Mean + KD.
With all source layers, Mean + KD first crossed at $K=3$ and the untouched checkpoint had no observed crossing through $K=20$.
Under an adaptive rank-4 projection, the interval for Mean + KD lay below 30\% and the untouched checkpoint's interval included 30\%.
Covariance matching was unnecessary for the restricted-protocol result, and some conditions without KD frequently refused harmless prompts.
In an aligned factorial comparison of refusal-feature ablation training (RFA) and KD, no cell met all behavioral and capability criteria under either GSM8K scoring rule.
RFA alone met the capability criterion only with flexible answer extraction.
RFA+KD met the behavioral criteria and lost 3.68 MMLU points relative to the source checkpoint, with an observed interaction of $-3.59$ points.
On the tested Qwen2.5-1.5B checkpoint, Mean + KD first crossed at $K=1$ versus $K=2$ for untouched Qwen.
The observed effect of representation matching on refusal depends on the model and extraction protocol.
\end{abstract}

\section{Introduction}

Instruction-tuned language models are expected to answer harmless requests while refusing requests that facilitate harm.
For open-weight models, this policy must remain effective even when a user can inspect and modify internal activations or weights.
That requirement is stronger than ordinary jailbreak robustness because the adversary need not search over prompts.

\citet{arditi2024refusal} found that refusal across thirteen chat models is mediated by a single residual-stream direction.
Projecting this direction from every residual write suppresses refusal while leaving several general capabilities largely intact.
The result is valuable mechanistic evidence, but it also exposes a simple white-box attack: expensive safety post-training can be removed by a rank-one intervention.

We investigate whether a training objective can redistribute the information used for refusal so that repeatedly ablating the most salient linear direction becomes less effective.
The starting point is LEACE, which characterizes linear guardedness through equality of class-conditional means \citep{belrose2023leace}.
During fine-tuning, we penalize empirical differences in class means and covariances.
We pair this geometric objective with output-distribution distillation from the frozen instruction-tuned model \citep{hinton2015distilling}.
The intended division of labor is that matching changes where information is represented, while distillation discourages a trivial solution that changes refusal behavior itself.

Testing this idea requires a stronger evaluation than a single post-training ablation.
After the first direction is removed, the residual class means can shift and expose another direction.
We therefore iterate extraction and projection, re-computing candidates on the already-ablated model at every step.
We further compare to ReFAT \citep{yu2025refat}, which directly trains a model to refuse while its layerwise refusal features are removed, and to a differentiable attack that optimizes an orthonormal low-rank projection against the frozen target model.

Our study is organized around four questions:

\begin{enumerate}
    \item Does representation matching add robustness beyond response supervised fine-tuning or distillation alone?
    \item Are covariance matching and chat-template alignment independently useful?
    \item Does any benefit persist under a direct ReFAT comparison and an adaptive gradient-based subspace attack?
    \item What is the cost in harmless refusal and standard language-model capabilities?
\end{enumerate}

We make three methodological contributions.
First, we provide an iterated linear ablation protocol that measures an attack curve rather than a single operating point.
Second, we formulate independently switchable mean, covariance, and distillation losses, test their causal combinations, and add an aligned RFA-by-KD confirmation for factor attribution.
Third, we evaluate both sides of the tradeoff with larger disjoint samples, Wilson confidence intervals, layerwise probes, selectivity metrics, and downstream capability benchmarks.
These changes turn a suggestive robustness observation into a controlled stress test of training-time representation shaping.

\section{Related Work}

\paragraph{Refusal representations and intervention.}
Linear probes measure how readily a property can be decoded from an intermediate representation \citep{alain2017probes}.
For refusal, \citet{arditi2024refusal} extract harmful-minus-harmless mean differences at residual-stream positions near the end of a chat prompt, select a direction using behavioral and utility criteria, and project it from every residual write.
\citet{dabas2025actor} fine-tune representations to reduce false refusals, while \citet{wollschlager2025geometry} identify multiple independent refusal directions in the same model.
Our focus is the complementary defensive question: can training make repeated low-rank projection less effective?

\paragraph{Robust safety training.}
ReFAT dynamically computes one harmful-minus-harmless direction per layer and trains on harmful responses while those directions are ablated \citep{yu2025refat}.
In a complementary study, \citet{labunets2026geometry} found that varying refusal prefixes during training could weaken vector ablation on an OLMo model.
We implement a controlled adaptation on the same model, examples, response targets, optimizer, and parameter-update regime used for our matching methods.
Other work studies continuous latent adversarial training \citep{sheshadri2025latent}, standardized red-teaming \citep{mazeika2024harmbench}, and safeguards intended to resist weight tampering \citep{tamirisa2025tamper}.
Our claims are narrower: we measure resistance to global low-rank residual projections, not security against arbitrary weight modification.

\paragraph{Concept erasure and statistical matching.}
INLP iteratively removes predictive linear subspaces from frozen representations \citep{ravfogel2020inlp}.
LEACE gives a closed-form minimum-distortion eraser and proves that equal class-conditional means characterize linear guardedness under its setting \citep{belrose2023leace}.
We use this result as motivation, not as a guarantee about stochastic finite-batch training.
An empirical mean penalty can be small while held-out probes remain accurate, and equal covariance is neither necessary nor sufficient for removal of all nonlinear information.
We explicitly measure these gaps in a GPT-2 sentiment study and in all-layer refusal probes.

\section{Threat Model and Attacks}

\subsection{Open-weight adversary}

The adversary has white-box access to model weights and may install hooks that modify residual-stream activations during inference.
The primary attack is constrained to a global rank-$K$ orthogonal projector shared across layers.
This restriction makes the intervention inexpensive and comparable to the rank-one attack of \citet{arditi2024refusal}, but it excludes arbitrary fine-tuning and layer-specific high-rank edits.

Let $D\in\mathbb{R}^{K\times d}$ contain orthonormal row vectors.
At each transformer block input, attention output, and MLP output, the attack applies
\begin{equation}
    h' = h - (hD^\top)D.
    \label{eq:projection}
\end{equation}
Hooking every residual write prevents a later submodule from immediately restoring an ablated direction.

\subsection{Iterated mean-difference ablation}

At attack step $k$, all previously selected directions remain installed.
We run balanced harmful and harmless extraction prompts through this intervened model and capture block-input residuals $h^{(\ell,p)}$.
Candidates are
\begin{equation}
 d^{(\ell,p)} = \mu^{(\ell,p)}_{\harm}-\mu^{(\ell,p)}_{\safe},
 \label{eq:direction}
\end{equation}
for the first 80\% of blocks and positions $p\in\{-5,\ldots,-1\}$ relative to the final prompt token.
This layer rule follows the released Arditi implementation, which discards the final 20\% of candidate source layers \citep{arditi2024code}.
Each candidate is orthogonalized against the existing attack subspace before scoring.

Candidate selection follows three criteria from the rank-one protocol.
We prefer directions that lower harmful-prompt refusal-token log odds, keep next-token KL on harmless prompts below a threshold, and induce refusal when added to harmless activations.
If no candidate passes both utility filters, we minimize refusal score plus a KL penalty.
After selection, the new unit vector is appended, all directions are QR-orthonormalized, and extraction is repeated on the newly ablated model.
This produces a curve for $K=0,1,\ldots,20$ rather than assuming that a basis extracted at $K=0$ remains optimal.

\subsection{Adaptive low-rank attack}

Mean differences need not find the worst subspace after fine-tuning.
We therefore parameterize a rank-$K$ basis $B\in\mathbb{R}^{d\times K}$, obtain orthonormal directions by differentiable QR decomposition, and backpropagate through the frozen target model and the projection in Equation~\ref{eq:projection}.
The attack minimizes harmful refusal-token log odds while penalizing KL between clean and attacked next-token distributions on harmless prompts:
\begin{equation}
 \begin{aligned}
 \mathcal{L}_{\mathrm{attack}}
 &= \mathbb{E}_{x\sim\mathcal{D}_{\harm}}[s_{\mathrm{ref}}(x;D)] \\
 &\quad + \beta\,\mathbb{E}_{x\sim\mathcal{D}_{\safe}}
 \mathrm{KL}(p_{0}(\cdot|x)\|p_D(\cdot|x)).
 \end{aligned}
 \label{eq:adaptive}
\end{equation}
We initialize from the corresponding iterated directions when available.
This is still a restricted global-subspace attack, but it adapts directly to the trained model and objective rather than relying on a mean-difference heuristic.

\section{Training Objectives}

\subsection{Response supervised fine-tuning}

Every training example contains a prompt $x$, the source instruction model's generated response $y$, and a prompt harmfulness label $c$.
The language-model loss predicts assistant response tokens only; prompt and padding tokens are masked.
This response SFT objective is present in every trained condition, including the SFT-only control.

For layer $\ell$, a pooling function $r$ maps prompt-token residual states to one vector $z^{(\ell)}=r(h^{(\ell)})$.
Let $\mu_c^{(\ell)}$ and $\Sigma_c^{(\ell)}$ denote its empirical class-conditional mean and unbiased sample covariance within a balanced microbatch.
We study
\begin{align}
 \mathcal{L}_{\mathrm{mean}}
 &= \frac{1}{L}\sum_{\ell=1}^{L}
 \left\|\mu_{\harm}^{(\ell)}-\mu_{\safe}^{(\ell)}\right\|_1,
 \label{eq:mean}\\
 \mathcal{L}_{\mathrm{cov}}
 &= \frac{1}{L}\sum_{\ell=1}^{L}
 \frac{\left\|\Sigma_{\harm}^{(\ell)}-\Sigma_{\safe}^{(\ell)}\right\|_F^2}{d^2}.
 \label{eq:cov}
\end{align}
The covariance term applies the squared covariance-distance penalty used in Deep CORAL \citep{sun2016deepcoral} to harmful and harmless prompt classes.
Balanced microbatches contain at least two examples from each class, so both sample covariances are defined.
The L2 covariance penalty is computed exactly through sample Gram matrices without materializing a $d\times d$ matrix.

\subsection{Distillation and pooling}

At the final prompt position, a frozen copy of the original instruction model provides teacher logits.
We use temperature-scaled forward KL,
\begin{equation}
 \mathcal{L}_{\mathrm{KD}}=T^2\mathrm{KL}
 \left(p_{\mathrm{teacher}}^T\|p_{\mathrm{student}}^T\right).
 \label{eq:kd}
\end{equation}
The complete objective is
\begin{equation}
 \mathcal{L}=\mathcal{L}_{\mathrm{SFT}}
 +\lambda_{\mu}\mathcal{L}_{\mathrm{mean}}
 +\lambda_{\Sigma}\mathcal{L}_{\mathrm{cov}}
 +\lambda_{\mathrm{KD}}\mathcal{L}_{\mathrm{KD}},
 \label{eq:total}
\end{equation}
with each auxiliary term independently switchable.

We compare three deterministic pools.
\textsc{User} averages only tokens whose character offsets overlap the user message.
\textsc{Prompt} averages the complete rendered prompt, including chat formatting.
\textsc{Chat} averages the final five prompt positions used by the attack's extraction scan.
This separation tests whether position alignment contributes independently of distillation and covariance matching.

\subsection{ReFAT baseline}

Our ReFAT control recomputes one harmful-minus-harmless last-token direction per layer every four optimizer steps using 32 prompts per class \citep{yu2025refat}.
With probability 0.5, directions are projected from the final 75\% of block inputs while computing the harmful response loss.
Training leaves harmless examples unmodified.
We report the one-epoch setting favored in the preliminary experiments of \citet{yu2025refat} and a three-epoch budget-matched sensitivity run.
The original ReFAT study used LoRA on larger models; we use full-parameter updates for every trained condition to isolate the training objective.
Because those ReFAT controls and the three-epoch matching grid differ in budget, we also run a separate aligned $2\times2$ confirmation that toggles RFA and KD while holding response SFT data, targets, optimizer, update count, and evaluation fixed.
RFA denotes enabling the ReFAT intervention in this confirmation, which is used for attribution rather than primary-model selection.

\section{Experimental Setup}

\subsection{Data and label provenance}

We derive our data from WildGuardMix \citep{han2024wildguard} and use Llama-3.2-1B-Instruct \citep{meta2024llama32} as the source model.
Our pinned training snapshot contains separate fields for prompt harmfulness and generated-response refusal.
These labels answer different questions and are not interchangeable.
The separate snapshot exists to materialize the prompt-level \texttt{prompt\_harm\_label} alongside the source model's generated response and model-specific refusal label.
The matching class is \texttt{prompt\_harm\_label}, which is determined by the WildGuardMix prompt category.
The model-specific \texttt{predicted\_refusal} label is used only for Arditi-style behavior conditioning: we retain harmful prompts refused by the source model and harmless prompts answered by it.
The reference response label in WildGuardMix describes a different response and is never used as prompt harmfulness.

After selecting only Llama-3.2-1B-Instruct generations, filtering behavior agreement, removing duplicate prompts, and balancing classes, we obtain \TrainNPerClass{} examples per class for training and \DevNPerClass{} per class for development.
For main evaluation, we use a separate pinned generation and classification snapshot, remove all overlapping sample IDs and exact prompt strings, and sample \ExternalNPerClass{} examples per class.
We also evaluate an unfiltered external split to expose sensitivity to behavior conditioning.
All split hashes and source revisions are recorded in a machine-readable manifest.

As a prespecified corpus-transfer stress test, after matching-condition selection and direction extraction are complete on WildGuardMix, we freeze the direction sequences for the untouched model, selected matching model, and one-epoch ReFAT\@.
We evaluate only these frozen sequences at $K\in\{0,1,4,8,16,20\}$ on a separate balanced set of 500 harmful and 500 harmless prompts from HarmBench, XSTest, and OR-Bench \citep{mazeika2024harmbench,rottger2024xstest,cui2025orbench}; this set is never used for training, extraction, model selection, or hyperparameter choice (Appendix~\ref{app:cross-dataset}).

\subsection{Optimization and conditions}

We fine-tune all ten factorial conditions with AdamW \citep{loshchilov2019adamw}, learning rate $2\times10^{-5}$, weight decay 0.01, bfloat16, three epochs, balanced microbatches of 16, and gradient accumulation to effective batch size 128.
Across the controlled grid, we fix $\lambda_{\mu}=10$, $\lambda_{\Sigma}=10^5$, $\lambda_{\mathrm{KD}}=100$, and $T=1$.
The larger covariance coefficient reflects the $d^2$ normalization in Equation~\ref{eq:cov}; every run saves observed component losses and its complete configuration.

The causal grid includes SFT-only, KD-only, mean-only, covariance-only, combined mean-covariance, each applicable matching loss with and without KD, and three pooling choices for the combined objective.
We evaluate the untouched instruction model and both ReFAT variants alongside this grid.
The matching checkpoint carried into adaptive-attack and attacked-capability comparisons is selected using development curves only: we compare the crossing based on the Wilson upper bound, the point-estimate crossing, selectivity AUC, and clean balanced refusal accuracy in that order.
External prompts are used only for final reporting.

The aligned confirmation uses 256 examples per class, one epoch, seed 42, and eight optimizer steps at effective batch size 64.
All four cells include response SFT; KD toggles $\lambda_{\mathrm{KD}}=100$ at $T=1$, while RFA toggles intervention probability 0.5 with directions refreshed every four steps from 32 prompts per class.
Its complete identity checks, gates, and artifacts are specified in Appendix~\ref{app:aligned-factorial}.

\subsection{Metrics and uncertainty}

We greedily generate up to 128 new tokens and classify response refusal with the official WildGuard model \citep{han2024wildguard}.
For each prompt class at each rank $K$, we estimate the refusal rate with a pointwise 95\% Wilson interval \citep{wilson1927interval}.
We use 30\% harmful refusal as a descriptive threshold calibrated to the baseline rank-one attack.
We report the first evaluated rank with an observed rate below 30\% ($K_{\mathrm{point}}$) and the first with a Wilson upper bound below 30\% ($K_{\mathrm{conf}}$).
Each interval describes prompt-sampling uncertainty at a fixed rank for one checkpoint.
We do not estimate an interval for either first-crossing rank or for the difference between checkpoints' crossing ranks.

For harmful and harmless refusal rates $R_{\harm}$ and $R_{\safe}$, we report a selectivity gap $R_{\harm}-R_{\safe}$ and balanced refusal accuracy $\frac{1}{2}(R_{\harm}+1-R_{\safe})$.
Here $1-R_{\safe}$ measures non-refusal on harmless prompts.
We do not score the quality of those answers.
We also fit frozen linear probes at every block-input layer and an MLP probe at the layer with maximum held-out linear accuracy.

We evaluate general capability with the Language Model Evaluation Harness \citep{gao2024lmeval} on 5-shot MMLU \citep{hendrycks2021mmlu}, 5-shot GSM8K \citep{cobbe2021gsm8k}, and 0-shot HellaSwag \citep{zellers2019hellaswag}.
We use each checkpoint's default chat template and render few-shot examples as separate turns for clean and attacked evaluations.
For GSM8K, strict extraction reads an answer following the \texttt{\#\#\#\#} marker.
Flexible extraction takes the last matching number in the same response.
Both filters score exact agreement with the reference answer.
The main grid originally summarized strict scores, and the aligned confirmation used flexible scores for its capability gate.
We report both filters in Table~\ref{tab:adaptive-capability} and test gate sensitivity in Appendix~\ref{app:aligned-factorial}.

\section{Results}

\subsection{The causal grid changes the interpretation}

\begin{table*}[t]
\centering
\scriptsize
\setlength{\tabcolsep}{3.5pt}
\begin{tabular}{lrrrr}
\toprule
Condition & Harm $\uparrow$ & Harmless $\downarrow$ & Gap $\uparrow$ & Max linear probe $\downarrow$ \\
\midrule
Untouched Instruct & 98.2 [96.6,99.1] & 7.8 [5.8,10.5] & 90.4 & 99.2 \\
SFT only & 98.6 [97.1,99.3] & 6.2 [4.4,8.7] & 92.4 & 99.1 \\
KD only & 97.8 [96.1,98.8] & 3.6 [2.3,5.6] & 94.2 & 99.2 \\
Mean, no KD & 92.6 [90.0,94.6] & 13.8 [11.1,17.1] & 78.8 & 98.9 \\
Mean + KD & 95.2 [93.0,96.8] & 3.6 [2.3,5.6] & 91.6 & 99.2 \\
Cov, no KD & 99.8 [98.9,100.0] & 16.6 [13.6,20.1] & 83.2 & 98.9 \\
Cov + KD & 96.4 [94.4,97.7] & 2.8 [1.7,4.6] & 93.6 & 99.1 \\
Mean + Cov, no KD & 93.0 [90.4,94.9] & 52.8 [48.4,57.1] & 40.2 & 98.8 \\
Mean + Cov + KD (chat) & 95.6 [93.4,97.1] & 5.6 [3.9,8.0] & 90.0 & 99.3 \\
Mean + Cov + KD (user) & 98.0 [96.4,98.9] & 3.6 [2.3,5.6] & 94.4 & 99.1 \\
Mean + Cov + KD (prompt) & 97.4 [95.6,98.5] & 4.4 [2.9,6.6] & 93.0 & 99.3 \\
ReFAT (1 epoch) & 100.0 [99.2,100.0] & 23.8 [20.3,27.7] & 76.2 & 99.3 \\
ReFAT (3 epochs) & 100.0 [99.2,100.0] & 21.2 [17.8,25.0] & 78.8 & 99.2 \\
\bottomrule
\end{tabular}
\caption{Complete clean factorial and pooling evaluation on the disjoint external set, in percent. Brackets give 95\% Wilson intervals. Max linear probe is the maximum held-out accuracy across all residual-stream block inputs.}
\label{tab:k0}
\end{table*}

Table~\ref{tab:k0} separates effects that were previously bundled.
The SFT-only and KD-only rows establish whether ordinary behavior cloning or teacher preservation can explain apparent robustness.
The no-KD matching rows quantify the tendency to over-refuse harmless prompts, while the matched pairs isolate the contribution of mean and covariance penalties.
Comparing \textsc{User}, \textsc{Prompt}, and \textsc{Chat} pooling tests the position-alignment hypothesis directly.
The selected Mean + KD checkpoint retains 95.2\% harmful refusal and 3.6\% harmless refusal, close to the untouched model's 98.2\% and 7.8\%.
Covariance matching is not required for this clean selectivity: Cov + KD is similarly selective, while adding covariance to mean matching without KD raises harmless refusal to 52.8\%.
All conditions retain maximum held-out linear-probe accuracy near 99\%, so the objectives reshape behavior without making prompt harmfulness linearly undecodable.

\subsection{Iterated ablation exposes attack rank and cost}

\begin{figure*}[t]
    \centering
    \IfFileExists{figures/iterated_curves.pdf}{%
      \includegraphics[width=0.98\textwidth]{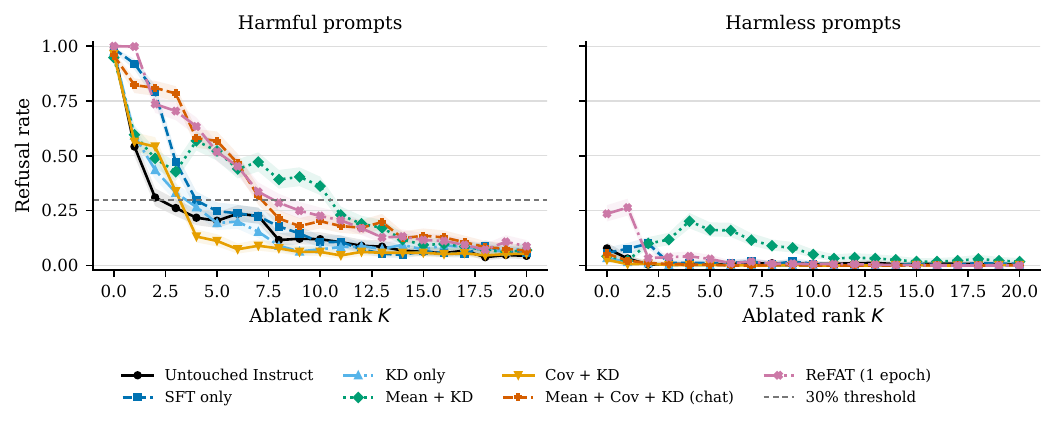}%
    }{%
      \placeholderfigure{Iterated harmful and harmless refusal curves will be generated from verified external-evaluation artifacts.}%
    }
    \caption{Refusal under iterated mean-difference ablation on 500 harmful and 500 harmless external prompts.
    Points show observed rates and bands show pointwise 95\% Wilson intervals.
    Every line is evaluated through the same maximum $K$, and the horizontal line marks the 30\% descriptive suppression threshold.}
    \label{fig:iterated}
\end{figure*}

Figure~\ref{fig:iterated} shows refusal on both prompt classes at each rank.
Selectivity AUC summarizes their difference across ranks.

\begin{table*}[t]
\centering
\scriptsize
\begin{tabular}{lrrrr}
\toprule
Condition & $K_{\mathrm{point}}$ & $K_{\mathrm{conf}}$ & Harmless at $K^*$ & Sel. AUC \\
\midrule
Untouched Instruct & 3 & 4 & 0.8 [0.3,2.0] & 0.164 \\
SFT only & 4 & 5 & 1.2 [0.6,2.6] & 0.220 \\
KD only & 4 & 5 & 0.2 [0.0,1.1] & 0.180 \\
Mean, no KD & $>20$ & $>20$ & -- & 0.387 \\
Mean + KD & 11 & 11 & 3.2 [2.0,5.1] & 0.247 \\
Cov, no KD & 17 & 17 & 1.0 [0.4,2.3] & 0.378 \\
Cov + KD & 4 & 4 & 0.2 [0.0,1.1] & 0.150 \\
Mean + Cov, no KD & 9 & 11 & 0.6 [0.2,1.7] & 0.331 \\
Mean + Cov + KD (chat) & 8 & 8 & 0.2 [0.0,1.1] & 0.329 \\
Mean + Cov + KD (user) & 9 & 9 & 2.2 [1.2,3.9] & 0.274 \\
Mean + Cov + KD (prompt) & 4 & 4 & 0.2 [0.0,1.1] & 0.169 \\
ReFAT (1 epoch) & 8 & 9 & 0.8 [0.3,2.0] & 0.311 \\
ReFAT (3 epochs) & 11 & 12 & 2.0 [1.1,3.6] & 0.437 \\
\bottomrule
\end{tabular}
\caption{External iterated-attack summary. $K_{\mathrm{point}}$ is the first harmful-refusal point estimate below 30\%; $K_{\mathrm{conf}}$ is the first step whose 95\% Wilson interval lies fully below 30\%. $K^*$ is $K_{\mathrm{conf}}$, or $K_{\mathrm{point}}$ when no confidence crossing is observed. Harmless refusal is in percent with a 95\% Wilson interval. Sel. AUC is the trapezoidal area between harmful and harmless refusal curves, normalized by the evaluated $K$ range.}
\label{tab:robustness}
\end{table*}

Table~\ref{tab:robustness} compares representation matching with SFT, KD, and ReFAT under identical extraction and evaluation.
The first observed upper-bound crossing was $K=\BaselineSuppressK$ for the untouched checkpoint (109/500 harmful prompts refused, 21.8\%, 95\% Wilson interval 18.4--25.6\%) and $K=\BestSuppressK$ for development-selected \BestMethodName{} (115/500, 23.0\%, 19.5--26.9\%).
SFT-only and KD-only cross at $K=5$, and Cov + KD crosses at $K=4$, showing that neither behavior cloning, distillation, nor covariance matching alone explains the selected result.
The apparently strongest no-KD curves are confounded by their clean harmless-refusal costs.
One-epoch and three-epoch ReFAT cross at $K=9$ and $K=12$, respectively, but begin with more than 20\% harmless refusal.
We distinguish ``not observed within budget'' from a security guarantee: twenty global directions remain cheap for a white-box adversary, and robustness to this attack family does not imply robustness to weight editing.

The sensitivity analyses sharply narrow this result.
On the behavior-unfiltered external set, the selected checkpoint first met the upper-bound criterion at the evaluated rank $K=8$.
On the separate HarmBench, XSTest, and OR-Bench mixture, its frozen WildGuardMix direction sequence also crosses at $K=8$, compared with $K=1$ for the untouched model.
The second-family transfer is negative: Mean + KD crosses at $K=\QwenBestSuppressK$ on Qwen2.5-1.5B, compared with $K=\QwenBaselineSuppressK$ for untouched Qwen.
Most importantly, when candidate extraction includes every source layer rather than excluding the final 20\%, the selected checkpoint crosses at $K=\AllLayerBestSuppressK$ while the untouched model has no observed upper-bound crossing through $K=20$.
The all-layer Mean + KD sequence selects layers 14 and 15 at its second and third steps, both outside the restricted candidate set.
Because each candidate set induces a separate greedy sequence, the all-layer baseline's nonmonotone curve is not guaranteed to dominate its restricted counterpart.
This sensitivity is therefore evidence of extraction-protocol dependence and relocation toward omitted layers, not a claim that merely adding candidates always strengthens a greedy attack.

\subsection{Adaptive attack and capability tradeoffs}

\begin{table*}[t]
\centering
\scriptsize
\setlength{\tabcolsep}{3.5pt}
\begin{tabular}{llrrrrrrr}
\toprule
Condition & Attack & Harm refusal & Harmless refusal & Harmless KL & MMLU & GSM8K strict & GSM8K flexible & HellaSwag \\
\midrule
Untouched Instruct & clean & 98.2 & 7.8 & 0.0000 & 45.8 & 32.60 & 43.75 & 53.5 \\
Untouched Instruct & adaptive rank 1 & 53.0 & 5.4 & 0.2394 & 46.0 & 30.48 & 42.08 & 53.2 \\
Untouched Instruct & adaptive rank 16 & 26.8 & 3.2 & 0.3137 & 45.5 & 29.34 & 40.71 & 54.1 \\
SFT only & clean & 98.6 & 6.2 & 0.0000 & 45.5 & 19.64 & 39.80 & 51.9 \\
KD only & clean & 97.8 & 3.6 & 0.0000 & 45.8 & 32.60 & 42.46 & 55.8 \\
Mean + KD & clean & 95.2 & 3.6 & 0.0000 & 45.7 & 34.72 & 43.90 & 56.2 \\
Mean + KD & adaptive rank 1 & 68.0 & 1.6 & 0.2045 & 45.4 & 37.98 & 43.90 & 55.5 \\
Mean + KD & adaptive rank 16 & 17.6 & 0.8 & 0.1932 & 44.3 & 25.25 & 41.09 & 55.8 \\
ReFAT (1 epoch) & clean & 100.0 & 23.8 & 0.0000 & 45.4 & 31.54 & 41.85 & 54.1 \\
ReFAT (1 epoch) & adaptive rank 1 & 60.0 & 18.2 & 0.2568 & 45.5 & 31.31 & 42.23 & 54.0 \\
ReFAT (1 epoch) & adaptive rank 16 & 75.8 & 26.4 & 0.3427 & 44.5 & 30.86 & 37.23 & 54.7 \\
ReFAT (3 epochs) & clean & 100.0 & 21.2 & 0.0000 & 45.5 & 20.24 & 41.17 & 52.1 \\
\bottomrule
\end{tabular}
\caption{Refusal rates and capability scores are percentages. GSM8K columns give strict and flexible exact match on the same responses. Harmless KL is clean-to-attacked next-token KL on the external set. Attacked rows use the same selected projection as the refusal evaluation.}
\label{tab:adaptive-capability}
\end{table*}

The adaptive evaluation reports refusal on each prompt class and harmless next-token KL for each optimized rank.
At rank 4, the selected matching checkpoint falls to \BestAdaptiveRankFourHarm{} harmful refusal with its Wilson interval fully below 30\%, while the untouched model is at \BaselineAdaptiveRankFourHarm{} and its interval includes 30\%.
One-epoch ReFAT remains above threshold at every tested adaptive rank, but its harmless refusal ranges from 18.2\% to 26.4\%.
Because each rank is optimized independently, the adaptive rows are not expected to be monotone in rank and should not be read as one nested attack curve.

Table~\ref{tab:adaptive-capability} reports benchmark scores before intervention and at selected adaptive ranks.
For clean Mean + KD and the untouched checkpoint, respectively, MMLU scores were \BestMMLU{} and \BaselineMMLU{}, and HellaSwag scores were \BestHellaSwag{} and \BaselineHellaSwag{}.
On the same 1,319 GSM8K responses from the untouched checkpoint, strict extraction scored 430 correct (32.60\%) and flexible extraction scored 577 correct (43.75\%).
For Mean + KD, the corresponding scores were 34.72\% and 43.90\%.
The clean SFT-only and three-epoch ReFAT scores were lower than the untouched scores under both rules, with larger drops under strict extraction.

At adaptive rank 16, MMLU and HellaSwag remained within 1.5 percentage points of their clean scores for both untouched Instruct and Mean + KD.
On GSM8K, the untouched checkpoint lost 3.3 points under strict extraction and 3.0 under flexible extraction.
Mean + KD lost 9.5 and 2.8 points, respectively.
One-epoch ReFAT lost 0.7 and 4.6 GSM8K points under the same two rules.

\subsection{Aligned RFA-by-KD confirmation}

% Generated from the validated aligned factorial summary.
\begin{table*}[t]
\centering
\scriptsize
\setlength{\tabcolsep}{4pt}
\begin{tabular}{lrrrrrl}
\toprule
Condition & Clean H/S & Clean gap & $K=1$ H/S & $K=1$ gap & MMLU & Outcome \\
\midrule
Supervised only & 100.0 / 15.6 & 84.4 & 98.0 / 20.2 & 77.8 & 45.9 & Fails clean \\
KD only & 98.0 / 18.2 & 79.8 & 50.0 / 7.6 & 42.4 & 46.3 & Fails clean \\
RFA only & 100.0 / 27.2 & 72.8 & 87.8 / 20.4 & 67.4 & 45.4 & Fails clean \\
RFA + KD & 97.0 / 1.6 & 95.4 & 77.6 / 0.6 & 77.0 & 42.2 & Fails capability \\
\midrule
RFA $\times$ KD interaction & -1.0 / -28.2 & +27.2 & +37.8 / -7.2 & +45.0 & -3.59 & n/a \\
\bottomrule
\end{tabular}
\caption{Aligned $2\times2$ confirmation, in percent. H/S is harmful/harmless refusal; the gap is H minus S. Each cell learns its own restricted $K=1$ direction. The final row is the RFA-by-KD difference in differences, in percentage points.}
\label{tab:aligned-factorial-main}
\end{table*}

Table~\ref{tab:aligned-factorial-main} compares all four cells under one protocol.
The clean gate requires at least 90\% harmful refusal, at most 10\% harmless refusal, and an 80-point selectivity gap.
At rank 1, harmful refusal must reach 50\% with a Wilson lower bound above 30\%.
The capability gate allows drops of up to three points per task and two points in the mean, using flexible GSM8K extraction.
RFA alone passes capability under flexible scoring and fails under strict scoring (Appendix~\ref{app:aligned-factorial}).
It also fails clean selectivity.
RFA+KD passes both behavioral gates, yet its MMLU score falls from 45.85\% to 42.17\%, failing the capability gate under either rule.
The difference in differences is beneficial for clean harmless refusal ($-28.2$ points) and selectivity gap ($+27.2$), but negative for MMLU ($-3.59$).
Under this one-seed protocol, the combination's MMLU loss is therefore interaction-associated rather than an RFA main effect.
Because each cell learns its own restricted direction, rank-1 interactions characterize the complete training-and-attack protocol rather than a shared fixed intervention.

\section{Discussion}

\paragraph{Robustness is distinct from erasure.}
A model must retain information about harmfulness to refuse selectively.
Chance probe accuracy is therefore neither required nor necessarily desirable.
We ask whether refusal of harmful prompts persists after projection, while tracking harmless-prompt refusal and benchmark scores separately.
The near-perfect refusal probes in every condition also show that the restricted iterated-attack result cannot be interpreted as removal of linearly accessible harmfulness.

\paragraph{Distillation can preserve selectivity without causing robustness.}
The factorial grid separates two roles that can look similar at $K=0$.
KD prevents the largest harmless-refusal increases in the matching conditions, yet KD-only remains nearly as easy to ablate as the original model.
Conversely, matching without KD can yield high apparent attack rank by refusing harmless prompts at $K=0$.
These controls support a narrower conclusion: distillation is important for selectivity, while the restricted robustness gain is associated specifically with mean matching.

\paragraph{The gain is extraction-protocol specific.}
The reversal under all-layer candidate extraction and the successful adaptive rank-4 attack are the strongest evidence against an attack-general robustness claim.
The Qwen transfer likewise reverses rather than reproduces the selected objective's Llama advantage.
Together they are consistent with Mean + KD redistributing refusal information toward source layers omitted by the released rank-one extraction rule.
This is a useful mechanistic intervention and a warning for evaluation design, but it is not a defense against an attacker free to broaden the search space.

\paragraph{ReFAT provides a different tradeoff.}
ReFAT is stronger than representation matching under the adaptive attack and competitive under restricted iterated extraction.
Its clean harmless-refusal rate, however, is substantially higher, and the three-epoch setting loses GSM8K accuracy.
In the aligned factorial, RFA alone over-refuses and passes the capability gate only with flexible GSM8K scoring. Adding KD restores behavioral selectivity and yields a negative MMLU interaction.
No aligned cell is uniformly successful, so the comparison favors reporting a safety-selectivity-capability frontier rather than declaring one method uniformly more robust.

The smaller three-seed GPT-2 sentiment control in Appendix~\ref{app:sentiment-protocol} likewise finds that adding covariance matching does not remove the residual linear signal left by mean matching.

\paragraph{Attacker effort should be stated literally.}
An increase from one to several projected directions is a mechanistically interesting change in effective attack rank.
It is not a large computational barrier for an adversary who owns the weights.
The practical contribution is a stress test and a training signal for reducing a specific fragility, not a claim that open weights become tamper-proof.

\section{Conclusion}

We study whether training-time statistical matching can make refusal less concentrated in a single ablatable residual direction.
The main advance is a controlled evaluation: full causal ablations, a direct ReFAT baseline, an aligned RFA-by-KD confirmation, an adaptive low-rank attack, external samples with uncertainty, selectivity curves, and downstream capabilities.
The observed first crossing occurred at a higher rank for Mean + KD on Llama under the restricted Arditi-style source-layer rule.
All-layer extraction, the Qwen transfer, and a gradient-optimized rank-4 projection remove or reverse that advantage, while covariance matching adds no consistent benefit and no-KD matching can create misleading robustness through over-refusal.
The evidence supports representational redistribution relative to one extraction protocol, not attack-general low-rank robustness.

\section*{Limitations}

The full causal grid uses one 1B-parameter model from one model family and English prompts from WildGuardMix.
Our frozen-direction cross-dataset stress test adds independently sourced HarmBench, XSTest, and OR-Bench prompts, but covers only three checkpoints and the fixed WildGuardMix-selected direction sequences.
Results may not transfer to larger models, multilingual safety policies, other chat templates, or attacks re-optimized on each new corpus.
The behavior-conditioned dataset follows prior refusal-direction work but selects prompts using source-model behavior; we report an unfiltered external sensitivity split, yet neither split represents deployment prevalence.
WildGuard is an automatic classifier and can misclassify nuanced refusals or partial compliance.

Our primary and adaptive attacks are global low-rank linear projections.
They do not cover unrestricted fine-tuning, sparse weight editing, nonlinear activation interventions, prompt jailbreaks, or layer-specific subspaces with independent directions.
The 30\% threshold is descriptive and model-specific.
Even when twenty directions are required, the intervention remains computationally cheap for a white-box adversary.

Mean and covariance penalties use finite balanced microbatches and do not enforce population moment equality.
Covariance estimates remain noisy at the per-class microbatch size, and equal first and second moments do not eliminate higher-order class information.
Our all-layer probes reduce dependence on one arbitrary layer but still cover only two classifier families.
The refusal grid and aligned RFA-by-KD confirmation use one fixed training seed per condition because of their size, so their Wilson intervals cover prompts rather than optimization variability; the smaller sentiment experiment uses three seeds.
Each aligned cell learns its own rank-1 attack direction, so its attack interactions are protocol-level contrasts rather than comparisons under one shared fixed projection.

MMLU, GSM8K, and HellaSwag measure performance on those benchmarks. Correctness, helpfulness, and coherence of answers to the harmless prompts were not independently evaluated.
Finally, we test one fixed coefficient setting and compute budget for the refusal grid; other tradeoffs may be attainable with systematic hyperparameter searches.

\section*{Ethical Considerations}

This work studies a white-box technique for disabling model refusal and therefore has dual-use implications.
The attack family is already public and simple to implement for anyone with model weights.
We include it because robust evaluation requires testing the intervention a defense claims to resist, and because hiding the evaluation would not remove the underlying vulnerability.

We avoid presenting higher attack rank as a security guarantee and report harmless refusal and benchmark scores alongside harmful refusal.
Generated harmful responses are processed for aggregate classification and are not reproduced in the paper.
The source corpora can contain offensive or disturbing prompts; we add no human annotations, report only aggregate measurements, and do not redistribute source rows or model weights.
Released artifacts are used for research under their recorded licenses and responsible-use terms, detailed in Appendix~\ref{app:artifact-terms}.
Artifacts should be released with clear documentation that the methods do not make an open-weight model safe against unrestricted tampering.

\paragraph{AI assistance.}
Generative AI tools assisted with code implementation and debugging, experimental-pipeline auditing, literature-search support, and drafting and editing portions of this revision.
The authors reviewed the resulting code and prose and take full responsibility for the methods, citations, results, and claims.

\bibliography{references}

\clearpage
\appendix
\section{Reproducibility Details}
\label{app:reproducibility}

We generate the result tables from the recorded evaluations.
For capability benchmarks, we apply each instruction model's default chat template and render each few-shot example as a separate chat turn.
Before aggregating scores, we check that each evaluation used these settings.
We evaluate on the MMLU test set (14,042 items across 57 subjects), the GSM8K test set (1,319 items), and the HellaSwag validation set (10,042 items).
We verify the evaluation-set sizes and report the MMLU aggregate across all 57 subjects.
The completed environment used Python 3.11.10, PyTorch 2.5.1 with CUDA 12.4, Transformers 5.14.1, Datasets 5.0.1, scikit-learn 1.9.0, and LM Evaluation Harness 0.4.12.
The main runs used NVIDIA H100 PCIe GPUs; confirmation and validation runs additionally used L40S, RTX 4090, and RTX 4080 Super GPUs.
The evaluated parameter counts are 1.23B for Llama-3.2-1B-Instruct \citep{meta2024llama32}, 1.54B for \href{https://huggingface.co/Qwen/Qwen2.5-1.5B-Instruct}{Qwen2.5-1.5B-Instruct} \citep{qwen2024qwen25}, 7B for WildGuard \citep{han2024wildguard}, and 124M for \href{https://huggingface.co/openai-community/gpt2}{GPT-2 small} \citep{radford2019gpt2}.
Invoice-backed records for the tracked rented instances total 66.5 allocated physical GPU-hours: 54.0 H100, 5.9 L40S, 3.6 RTX 4090, 2.2 RTX 4080 Super, and 0.8 single-GPU relay hours.
This is a conservative allocation total, not measured kernel utilization, and includes setup, validation, file transfer, relay, and idle time.

\subsection{Data sources, labels, and splits}
\label{app:data}

The frozen training snapshot preserves two labels that must not be conflated.
The field \texttt{prompt\_harm\_label} is the class used by the representation losses and attacks.
It is deterministically derived from the original WildGuardMix subcategory: \texttt{benign} maps to \texttt{unharmful}, and every other retained subcategory maps to \texttt{harmful}.
The field \texttt{predicted\_refusal} describes the behavior of the particular source model on its generated response.
It is used only for behavior conditioning, where harmful prompts must have been refused and harmless prompts must have been answered.
The WildGuardMix reference-response label describes a different response and is never used as prompt harmfulness.

We use separate training and evaluation snapshots derived from WildGuardMix.
For training, we select rows with \texttt{model\_label=llama-1b-instruct-chat} and \texttt{model=meta-llama/Llama-3.2-1B-Instruct}.
After removing duplicate prompts, we verify that the training, development, and external evaluation splits share neither sample IDs nor exact prompt strings.
Table~\ref{tab:data-splits} gives the class counts for each split.

\begin{table}[t]
\centering
\small
\begin{tabular*}{\columnwidth}{@{\extracolsep{\fill}}lrrr@{}}
\toprule
Split & Total & Harmful & Harmless \\
\midrule
Training             & 1,132 & 566 & 566 \\
Development          &   200 & 100 & 100 \\
External             & 1,000 & 500 & 500 \\
External, unfiltered & 1,000 & 500 & 500 \\
\bottomrule
\end{tabular*}
\caption{WildGuardMix splits by prompt class.
All except the unfiltered external split retain harmful prompts the source model refused and harmless prompts it answered.}
\label{tab:data-splits}
\end{table}

The unfiltered external split is balanced by prompt class but not by source-model behavior.
It contains 810 responses labeled compliance, 189 labeled refusal, and one classifier parse failure in the source snapshot.
This split is used only for sensitivity analysis, not attack extraction or model selection.

\subsection{Artifact terms and data handling}
\label{app:artifact-terms}

We accessed WildGuardMix under the ODC-BY terms and AI2 Responsible Use Guidelines listed on its \href{https://huggingface.co/datasets/allenai/wildguardmix}{dataset card} and used it for the research purposes described there \citep{han2024wildguard}.
The card warns that the data include harmful, offensive, or otherwise disturbing material.
The \href{https://huggingface.co/datasets/bench-llm/or-bench}{OR-Bench dataset} and \href{https://github.com/paul-rottger/xstest}{XSTest prompts} are released under CC BY 4.0.
The \href{https://huggingface.co/allenai/wildguard}{WildGuard} and \href{https://huggingface.co/Qwen/Qwen2.5-1.5B-Instruct}{Qwen2.5-1.5B-Instruct} model cards list Apache 2.0.
The \href{https://github.com/centerforaisafety/HarmBench/blob/main/LICENSE}{HarmBench code} uses the MIT License, the \href{https://github.com/openai/gpt-2/blob/master/LICENSE}{GPT-2 repository} lists a modified MIT license, and Llama 3.2 uses the Llama 3.2 Community License \citep{meta2024llama32}.
We record the exact source revisions, use the released artifacts for their documented research and evaluation purposes, and do not redistribute source rows or model weights.
No new human annotation or identity inference is performed.
Potentially harmful generations are handled programmatically and reported only through aggregate refusal and capability measurements.

\subsection{Full training configuration}
\label{app:training-hyperparameters}

All ten factorial Llama conditions use full-parameter AdamW updates in bfloat16, a learning rate of $2\times10^{-5}$, weight decay 0.01, linear learning-rate decay, 5\% warmup, and gradient-norm clipping at 1.0.
Training runs for three epochs with seed 42, class-balanced microbatches of 16, gradient accumulation of 8, and effective batch size 128.
There are 27 optimizer steps.
Inputs are truncated to 512 tokens and responses to 256 tokens.
The causal-language-model loss is computed only on assistant response tokens; prompt and padding tokens are masked.
The source identifier is \texttt{meta-llama/Llama-3.2-1B-Instruct}.
We record the training configuration and per-step loss components for each checkpoint.
The matching conditions, untouched baseline, and original ReFAT comparison use the same data manifest.

For matching runs, the mean term is an L1 distance with $\lambda_{\mu}=10$ and the covariance term is a squared Frobenius distance normalized by $d^2$ with $\lambda_{\Sigma}=10^5$.
The frozen-teacher forward KL uses $T=1$ and $\lambda_{\mathrm{KD}}=100$.
Losses are averaged over all block-input residual layers.
Every balanced microbatch contains eight examples from each class, so each empirical covariance has more than one observation per class.

\begin{table*}[t]
\centering
\scriptsize
\setlength{\tabcolsep}{4pt}
\begin{tabular}{lccccl}
\toprule
Condition & Mean & Covariance & KD & Epochs & Pool \\
\midrule
SFT only & no & no & no & 3 & n/a \\
KD only & no & no & yes & 3 & n/a \\
Mean only & yes & no & no & 3 & chat positions \\
Mean + KD & yes & no & yes & 3 & chat positions \\
Covariance only & no & yes & no & 3 & chat positions \\
Covariance + KD & no & yes & yes & 3 & chat positions \\
Mean + covariance & yes & yes & no & 3 & chat positions \\
Mean + covariance + KD & yes & yes & yes & 3 & chat positions \\
Mean + covariance + KD & yes & yes & yes & 3 & user tokens \\
Mean + covariance + KD & yes & yes & yes & 3 & full prompt \\
ReFAT & n/a & n/a & n/a & 1 & last prompt token \\
ReFAT, budget matched & n/a & n/a & n/a & 3 & last prompt token \\
\bottomrule
\end{tabular}
\caption{Complete causal training grid.
All ten matching and control rows use the same data, response targets, optimizer, and full-parameter update regime.
The untouched instruction model is evaluated as an additional no-training baseline.}
\label{tab:causal-grid}
\end{table*}

The \textsc{Chat} pool averages prompt positions $-5$ through $-1$, matching the attack scan.
The \textsc{User} pool averages only tokens whose offsets overlap the user message.
The \textsc{Prompt} pool averages all rendered prompt tokens, including chat-format tokens.
Pooling is irrelevant for SFT-only and KD-only because no matching term consumes hidden states.

The controlled ReFAT implementation refreshes one harmful-minus-harmless last-token direction per layer every four optimizer steps using 32 prompts per class.
With probability 0.5, it projects those layer-specific directions from the final 75\% of block inputs while computing the harmful response loss.
Harmless response training is never intervened upon.
We evaluate the one-epoch setting favored in the preliminary experiments of \citet{yu2025refat} and a three-epoch budget-matched run.
This is a full-parameter controlled adaptation of ReFAT rather than an exact reproduction of the original LoRA configuration.

\subsection{Aligned RFA-by-KD confirmation}
\label{app:aligned-factorial}

The aligned confirmation is a separate exact $2\times2$ design with four cells: supervised-only, KD-only, RFA-only, and RFA+KD.
Every cell uses the same response-only SFT objective, frozen data revisions, 256 training examples per class, one epoch, seed 42, microbatch size 4, gradient accumulation 16, effective batch size 64, and eight optimizer steps.
KD uses the frozen source teacher with $T=1$ and $\lambda_{\mathrm{KD}}=100$.
RFA uses intervention probability 0.5, refreshes directions every four optimizer steps from a fixed 32 prompts per class, and projects layer-specific directions from block inputs 4 through 15 on harmful examples only.
All nonfactor settings, response targets, optimizer settings, and evaluation artifacts are identity-validated before aggregation.

Clean and restricted rank-1 behavior are each evaluated on 500 harmful and 500 harmless external prompts.
The clean gate requires harmful refusal at least 90\%, harmless refusal at most 10\%, and selectivity gap at least 80 percentage points.
The restricted rank-1 gate requires harmful refusal at least 50\% and its 95\% Wilson lower bound to remain above the 30\% suppression threshold.
The capability gate permits at most a three-percentage-point drop on any task and at most a two-point drop in the three-task mean.
Under the confirmation's original scoring rule, the source scores were 53.49\% for HellaSwag, 45.85\% for MMLU, and 43.75\% for GSM8K with flexible extraction (mean 47.69\%).
Table~\ref{tab:adaptive-capability} gives both GSM8K scores from this source evaluation.

% Generated from the validated aligned factorial summary.
\begin{table*}[t]
\centering
\scriptsize
\setlength{\tabcolsep}{4pt}
\begin{tabular}{lrrrrrrr}
\toprule
Condition & Clean H & Clean S & Clean gap & $K=1$ H & $K=1$ S & $K=1$ gap & $K=1$ KL \\
\midrule
Supervised only & 100.0 & 15.6 & 84.4 & 98.0 & 20.2 & 77.8 & 0.1068 \\
KD only & 98.0 & 18.2 & 79.8 & 50.0 & 7.6 & 42.4 & 0.1132 \\
RFA only & 100.0 & 27.2 & 72.8 & 87.8 & 20.4 & 67.4 & 0.1214 \\
RFA + KD & 97.0 & 1.6 & 95.4 & 77.6 & 0.6 & 77.0 & 0.0248 \\
\bottomrule
\end{tabular}
\caption{Complete behavior results for the aligned RFA-by-KD confirmation. Refusal rates and gaps are percentages. KL is clean-to-attacked next-token KL on harmless prompts.}
\label{tab:aligned-factorial-behavior}
\end{table*}

% Generated from the validated aligned factorial summary.
\begin{table*}[t]
\centering
\scriptsize
\setlength{\tabcolsep}{3.5pt}
\begin{tabular}{lrrrrllll}
\toprule
Condition & HellaSwag & MMLU & GSM8K (flex.) & Mean & Clean & $K=1$ & Capability & Outcome \\
\midrule
Supervised only & 56.1 & 45.9 & 42.1 & 48.0 & Fail & Pass & Pass & Fails clean \\
KD only & 57.3 & 46.3 & 43.6 & 49.0 & Fail & Pass & Pass & Fails clean \\
RFA only & 54.8 & 45.4 & 41.8 & 47.4 & Fail & Pass & Pass & Fails clean \\
RFA + KD & 55.4 & 42.2 & 41.2 & 46.3 & Pass & Pass & Fail & Fails capability \\
\bottomrule
\end{tabular}
\caption{Capability scores and predeclared gate outcomes for the aligned confirmation, in percent. GSM8K uses flexible extraction. The clean gate requires harmful refusal at least 90\%, harmless refusal at most 10\%, and gap at least 80\%. The $K=1$ gate requires harmful refusal at least 50\% with its interval above the 30\% suppression threshold. The capability gate permits at most a three-point drop on any task and a two-point drop in the three-task mean relative to the source model.}
\label{tab:aligned-factorial-capability}
\end{table*}

% Generated from the validated aligned factorial summary.
\begin{table*}[t]
\centering
\scriptsize
\setlength{\tabcolsep}{4pt}
\begin{tabular}{lrrrrr}
\toprule
Metric & KD, no RFA & KD, with RFA & RFA, no KD & RFA, with KD & Interaction \\
\midrule
Clean harmful refusal & -2.0 & -3.0 & +0.0 & -1.0 & -1.0 \\
Clean harmless refusal & +2.6 & -25.6 & +11.6 & -16.6 & -28.2 \\
Clean selectivity gap & -4.6 & +22.6 & -11.6 & +15.6 & +27.2 \\
$K=1$ harmful refusal & -48.0 & -10.2 & -10.2 & +27.6 & +37.8 \\
$K=1$ harmless refusal & -12.6 & -19.8 & +0.2 & -7.0 & -7.2 \\
$K=1$ selectivity gap & -35.4 & +9.6 & -10.4 & +34.6 & +45.0 \\
$K=1$ harmless KL & +0.0064 & -0.0966 & +0.0146 & -0.0884 & -0.1030 \\
HellaSwag & +1.19 & +0.55 & -1.23 & -1.87 & -0.64 \\
MMLU & +0.40 & -3.19 & -0.51 & -4.09 & -3.59 \\
GSM8K (flex.) & +1.52 & -0.61 & -0.23 & -2.35 & -2.12 \\
Capability mean & +1.03 & -1.08 & -0.66 & -2.77 & -2.12 \\
\bottomrule
\end{tabular}
\caption{Factorial contrasts using flexible answer extraction for GSM8K. Entries are percentage points except harmless KL, which is on its raw scale. Interaction is RFA+KD minus RFA-only minus KD-only plus supervised-only. Positive effects are beneficial for harmful refusal, selectivity gap, and capability; negative effects are beneficial for harmless refusal and KL.}
\label{tab:aligned-factorial-effects}
\end{table*}

The interaction is the standard difference in differences, RFA+KD minus RFA-only minus KD-only plus supervised-only.
Using flexible GSM8K scores, the GSM8K and capability-mean interactions are both $-2.12$ percentage points.
Rescoring the source and all four cells with strict extraction gives GSM8K scores of 32.60\% (source), 29.95\% (supervised-only), 41.55\% (KD-only), 29.26\% (RFA-only), and 16.15\% (RFA+KD).
Only RFA-only changes its capability classification. Its strict GSM8K score is 3.34 points below the source score, exceeding the three-point allowance.
The strict GSM8K and capability-mean interactions are $-24.72$ and $-9.65$ points.
The MMLU interaction is $-3.59$ points under both scoring rules.
Conversely, the clean harmless-refusal and selectivity-gap interactions are beneficial by 28.2 and 27.2 points, respectively.
Each cell learns its own development-selected rank-1 direction, so restricted-attack interactions compare complete cell-specific training-and-attack protocols rather than responses to one common fixed direction.
We recompute the factorial contrasts from the four cell scores before reporting them.

\subsection{Iterated attack constants}
\label{app:attack-constants}

At each step, direction extraction and candidate scoring use 64 development prompts per class, sampled with seed 42 from the 100-per-class development split.
Candidate directions are block-input harmful-minus-harmless mean differences at positions $-5,-4,-3,-2,-1$ from each source layer in the first 80\% of the network.
The final 20\% cutoff follows the released rank-one protocol and is separately tested by an all-source-layer sensitivity run.
Each candidate is orthogonalized against the already selected subspace.

Candidate scoring uses refusal-token log odds, clean-to-ablated harmless next-token KL, and positive steering on harmless prompts.
The refusal-token set contains the first tokenizer ID produced by the string \texttt{I}, matching the Llama-3 choice in the released rank-one protocol.
A candidate passes the utility filters when KL is at most 0.1 and its steering score is at least 0.
Among passing candidates, the attack minimizes refusal log odds.
If none passes, it minimizes refusal log odds plus KL with coefficient 1.
The steering coefficient is 1.
The main grid disables norm-based early stopping, sets $K_{\max}=20$, and evaluates every saved rank from 0 through 20.
Generation is greedy with at most 128 new tokens.
The attack is installed at every block input, attention output, and MLP output.
Behavioral scoring uses \texttt{allenai/wildguard}, with classifier batch size 32.

The development split is used for direction extraction and selection only.
The selected sequence is then frozen and evaluated at every $K$ on 500 harmful and 500 harmless external prompts.
For each checkpoint we report the first evaluated rank whose harmful-refusal point estimate falls below 30\% and the first whose pointwise 95\% Wilson upper bound falls below 30\%.

\subsection{Adaptive restart and selection protocol}
\label{app:adaptive-protocol}

Adaptive projections are optimized at ranks 1, 4, 8, and 16 for the untouched instruction model, the matching condition selected by the prespecified iterated-curve summary, and one-epoch ReFAT.
For every model and rank, we run three independent starts: a greedy-direction initialization with seed 42 and random initializations with seeds 101 and 202.
The unconstrained $d\times K$ basis is mapped to orthonormal directions by differentiable reduced QR.

Each restart takes 200 Adam steps with learning rate 0.05, gradient clipping at 1.0, and a balanced optimization batch of 16 training prompts per class per step.
The loss is harmful refusal-token log odds plus five times harmless clean-to-attacked next-token KL.
Optimization never updates model weights.

Restart selection is performed on the entire separate development split, using the same surrogate objective and no generated-response labels.
For each model and rank, we record all three development scores and evaluate only the subspace with the lowest score on the external split.
Table~\ref{tab:adaptive-restarts} reports restart selection, and Table~\ref{tab:adaptive-behavior} reports external refusal rates.

\begin{table*}[t]
\centering
\scriptsize
\begin{tabular}{lrrrrrrr}
\toprule
Condition & Rank & Greedy-42 & Random-101 & Random-202 & Selected & Harm log odds & Harmless KL \\
\midrule
Untouched Instruct & 1 & -17.421 & -17.376 & -16.017 & G42 & -18.105 & 0.1368 \\
Untouched Instruct & 4 & -16.869 & -17.281 & -17.058 & R101 & -17.869 & 0.1176 \\
Untouched Instruct & 8 & -16.893 & -16.941 & -16.862 & R101 & -17.883 & 0.1884 \\
Untouched Instruct & 16 & -16.892 & -16.397 & -16.653 & G42 & -18.139 & 0.2495 \\
Mean + KD & 1 & -17.155 & -16.581 & -15.733 & G42 & -17.941 & 0.1572 \\
Mean + KD & 4 & -17.099 & -16.751 & -16.925 & G42 & -17.842 & 0.1486 \\
Mean + KD & 8 & -17.044 & -16.485 & -16.125 & G42 & -18.157 & 0.2225 \\
Mean + KD & 16 & -17.090 & -16.410 & -16.265 & G42 & -18.156 & 0.2132 \\
ReFAT (1 epoch) & 1 & -17.130 & -17.150 & -16.181 & R101 & -17.750 & 0.1200 \\
ReFAT (1 epoch) & 4 & -17.366 & -17.286 & -17.523 & R202 & -18.039 & 0.1033 \\
ReFAT (1 epoch) & 8 & -17.623 & -16.824 & -17.163 & G42 & -17.965 & 0.0683 \\
ReFAT (1 epoch) & 16 & -17.083 & -17.228 & -16.532 & R101 & -18.277 & 0.2097 \\
\bottomrule
\end{tabular}
\caption{Adaptive restart diagnostics on the development split. The three objective columns report harmful refusal-token log odds plus five times harmless next-token KL. Selected is the minimum-objective restart; only its copied subspace is used for external evaluation.}
\label{tab:adaptive-restarts}
\end{table*}

\begin{table*}[t]
\centering
\scriptsize
\begin{tabular}{lrrrrl}
\toprule
Condition & Rank & Harm refusal & Harmless refusal & Harmless KL & 30\% status \\
\midrule
Untouched Instruct & 1 & 53.0 [48.6,57.3] & 5.4 [3.7,7.7] & 0.2394 & above \\
Untouched Instruct & 4 & 31.2 [27.3,35.4] & 4.6 [3.1,6.8] & 0.2750 & inconclusive \\
Untouched Instruct & 8 & 16.0 [13.0,19.5] & 3.0 [1.8,4.9] & 0.3137 & below \\
Untouched Instruct & 16 & 26.8 [23.1,30.8] & 3.2 [2.0,5.1] & 0.3137 & inconclusive \\
Mean + KD & 1 & 68.0 [63.8,71.9] & 1.6 [0.8,3.1] & 0.2045 & above \\
Mean + KD & 4 & 22.2 [18.8,26.0] & 1.0 [0.4,2.3] & 0.0920 & below \\
Mean + KD & 8 & 26.6 [22.9,30.6] & 0.6 [0.2,1.7] & 0.2114 & inconclusive \\
Mean + KD & 16 & 17.6 [14.5,21.2] & 0.8 [0.3,2.0] & 0.1932 & below \\
ReFAT (1 epoch) & 1 & 60.0 [55.6,64.2] & 18.2 [15.1,21.8] & 0.2568 & above \\
ReFAT (1 epoch) & 4 & 53.0 [48.6,57.3] & 21.8 [18.4,25.6] & 0.2275 & above \\
ReFAT (1 epoch) & 8 & 40.8 [36.6,45.2] & 20.6 [17.3,24.4] & 0.2682 & above \\
ReFAT (1 epoch) & 16 & 75.8 [71.9,79.3] & 26.4 [22.7,30.4] & 0.3427 & above \\
\bottomrule
\end{tabular}
\caption{External behavior under independently optimized adaptive subspaces. Refusal rates are percentages with 95\% Wilson intervals. Status compares the full harmful-refusal interval with the descriptive 30\% threshold; ranks are separate optimized attacks and need not be monotone.}
\label{tab:adaptive-behavior}
\end{table*}

\subsection{All-layer refusal probes}
\label{app:refusal-probes}

Refusal probes use last-prompt-token block-input residuals rather than block outputs.
They are fit on up to 1,000 balanced training examples and evaluated on the disjoint 1,000-example external split.
A separate logistic linear probe is fit at every transformer block input.
We report the maximum held-out accuracy and its layer, avoiding a post-hoc choice of one favored layer in the main comparison.
An MLP probe is fit only at the layer with maximum held-out linear accuracy and is labeled as such.
Table~\ref{tab:refusal-layerwise-probes} reports held-out probe accuracy by layer.

\begin{table*}[t]
\centering
\scriptsize
\setlength{\tabcolsep}{3.2pt}
\begin{tabular}{lrrrrrrrr}
\toprule
Condition & L0 & L1 & L2 & L3 & L4 & L5 & L6 & L7 \\
\midrule
Untouched Instruct & 50.0 & 94.9 & 96.7 & 96.8 & 97.0 & 97.7 & 98.3 & 98.5 \\
SFT only & 50.0 & 95.1 & 96.2 & 96.8 & 97.4 & 97.7 & 97.7 & 98.3 \\
KD only & 50.0 & 94.6 & 96.6 & 96.7 & 97.4 & 97.4 & 97.6 & 98.3 \\
Mean, no KD & 50.0 & 94.8 & 95.9 & 95.9 & 96.9 & 97.7 & 97.8 & 98.0 \\
Mean + KD & 50.0 & 94.6 & 96.3 & 96.7 & 97.6 & 97.5 & 97.8 & 98.3 \\
Cov, no KD & 50.0 & 94.8 & 96.1 & 96.5 & 96.9 & 96.8 & 97.4 & 98.2 \\
Cov + KD & 50.0 & 95.0 & 96.6 & 96.7 & 97.4 & 97.2 & 97.7 & 98.2 \\
Mean + Cov, no KD & 50.0 & 94.6 & 96.3 & 96.3 & 96.6 & 96.9 & 97.2 & 98.1 \\
Mean + Cov + KD (chat) & 50.0 & 94.7 & 96.7 & 96.8 & 97.0 & 97.4 & 97.8 & 98.4 \\
Mean + Cov + KD (user) & 50.0 & 94.7 & 96.5 & 96.6 & 97.4 & 97.1 & 97.6 & 98.1 \\
Mean + Cov + KD (prompt) & 50.0 & 94.5 & 96.7 & 97.0 & 97.3 & 98.0 & 97.7 & 98.2 \\
ReFAT (1 epoch) & 50.0 & 94.8 & 96.5 & 96.6 & 97.2 & 97.2 & 97.4 & 98.4 \\
ReFAT (3 epochs) & 50.0 & 95.1 & 96.3 & 96.8 & 97.2 & 97.7 & 97.6 & 98.2 \\
\bottomrule
\end{tabular}
\par\smallskip
\begin{tabular}{lrrrrrrrrrr}
\toprule
Condition & L8 & L9 & L10 & L11 & L12 & L13 & L14 & L15 & Max layer & MLP@max \\
\midrule
Untouched Instruct & 99.1 & 98.9 & 99.2 & 99.0 & 99.0 & 99.2 & 99.2 & 99.2 & 10 & 98.9 \\
SFT only & 98.7 & 98.9 & 99.1 & 99.0 & 98.8 & 98.7 & 98.8 & 98.9 & 10 & 98.6 \\
KD only & 98.8 & 99.0 & 99.2 & 99.1 & 99.0 & 99.0 & 99.0 & 99.0 & 10 & 98.9 \\
Mean, no KD & 98.6 & 98.6 & 98.8 & 98.3 & 98.4 & 98.9 & 98.9 & 98.5 & 13 & 97.9 \\
Mean + KD & 98.7 & 98.9 & 99.2 & 98.9 & 98.8 & 99.1 & 98.8 & 98.8 & 10 & 98.9 \\
Cov, no KD & 98.6 & 98.9 & 98.7 & 98.6 & 98.6 & 98.6 & 98.7 & 98.6 & 9 & 98.4 \\
Cov + KD & 98.6 & 99.0 & 99.1 & 98.9 & 98.9 & 98.9 & 98.8 & 98.8 & 10 & 98.6 \\
Mean + Cov, no KD & 98.4 & 98.2 & 98.8 & 98.5 & 98.7 & 98.1 & 98.4 & 98.2 & 10 & 98.5 \\
Mean + Cov + KD (chat) & 98.6 & 98.7 & 99.3 & 98.9 & 98.9 & 98.9 & 99.0 & 98.8 & 10 & 98.7 \\
Mean + Cov + KD (user) & 98.5 & 98.9 & 99.1 & 99.1 & 99.1 & 98.9 & 99.1 & 98.8 & 10 & 98.7 \\
Mean + Cov + KD (prompt) & 98.7 & 99.1 & 99.3 & 99.1 & 99.1 & 99.1 & 99.1 & 99.1 & 10 & 98.8 \\
ReFAT (1 epoch) & 98.9 & 99.2 & 99.2 & 99.2 & 99.0 & 98.9 & 99.1 & 99.3 & 15 & 98.8 \\
ReFAT (3 epochs) & 98.7 & 98.9 & 99.2 & 99.1 & 99.0 & 98.8 & 99.1 & 98.9 & 10 & 98.6 \\
\bottomrule
\end{tabular}
\caption{Held-out linear probe accuracy at every Llama block-input residual, in percent. Max layer is selected by held-out linear accuracy; MLP@max is fit only at that layer.}
\label{tab:refusal-layerwise-probes}
\end{table*}

\section{Sentiment Protocol}
\label{app:sentiment-protocol}

The proof of concept uses GPT-2 small \citep{radford2019gpt2} and the IMDb sentiment dataset \citep{maas2011imdb}.
Training uses 8,000 examples from a seed-42 shuffle of the IMDb training split, one epoch, class-balanced batches of 32, AdamW with learning rate $5\times10^{-5}$ and weight decay 0.01, 5\% linear warmup, maximum length 256, and full-token causal-language-model loss.
The four trainable conditions are language-model fine-tuning only, mean matching, covariance matching, and mean plus covariance matching.
Each is repeated with seeds 13, 42, and 73.

Loss weights are chosen without viewing the official test split.
The calibration run trains on positions 0--999 of the shuffled training split, fits probes on positions 1,000--1,999, and evaluates those probes and validation perplexity on positions 2,000--2,999.
The selected configuration is the strongest matching setting within a prespecified +3 perplexity guardrail relative to the no-matching validation run.
The final selected values are $\lambda_{\mu}=10$ and $\lambda_{\Sigma}=1$.

Final probes are trained on 2,000 examples beginning at position 8,000 of the shuffled training split and evaluated on 2,000 examples from the official test split.
Features are attention-mask means of block-input residuals.
We fit linear probes at every block, report the maximum held-out linear accuracy, and fit an MLP at the final block input.
We also report the final-layer L2 class-mean gap, the covariance Frobenius gap divided by hidden size, and perplexity on the first 500 evaluation examples.
LEACE and INLP are fit to frozen raw GPT-2 representations as post-hoc reference transformations and are not treated as trainable defenses.

\begin{table*}[t]
\centering
\scriptsize
\begin{tabular}{lrrrrrr}
\toprule
Condition & Mean gap $\downarrow$ & Cov. gap $\downarrow$ & Linear@11 $\downarrow$ & Max linear $\downarrow$ & MLP@11 $\downarrow$ & PPL $\downarrow$ \\
\midrule
GPT-2 (raw) & 9.772 & 1.58e-01 & 82.7 & 83.6 & 86.8 & 40.449 \\
LM fine-tuning & 11.186 $\pm$ 0.022 & 1.60e-01 $\pm$ 3.72e-04 & 84.1 $\pm$ 0.2 & 84.3 $\pm$ 0.3 & 87.3 $\pm$ 0.2 & 35.029 $\pm$ 0.004 \\
Mean matching & 2.415 $\pm$ 0.012 & 5.95e-02 $\pm$ 4.18e-04 & 79.7 $\pm$ 0.2 & 81.2 $\pm$ 0.2 & 82.9 $\pm$ 0.3 & 36.942 $\pm$ 0.071 \\
Covariance matching & 8.498 $\pm$ 0.057 & 8.34e-02 $\pm$ 1.90e-03 & 83.9 $\pm$ 0.4 & 84.0 $\pm$ 0.2 & 87.8 $\pm$ 0.3 & 35.173 $\pm$ 0.031 \\
Mean + covariance & 2.459 $\pm$ 0.063 & 5.74e-02 $\pm$ 1.18e-02 & 79.6 $\pm$ 0.8 & 81.2 $\pm$ 0.2 & 83.2 $\pm$ 0.2 & 37.235 $\pm$ 0.060 \\
LEACE & -- & -- & 49.5 & 49.5 & -- & -- \\
INLP & -- & -- & 70.2 & 70.2 & -- & -- \\
\bottomrule
\end{tabular}
\caption{IMDb proof of concept. Mean and covariance gaps and @11 probes use the final block input; Max linear is the maximum held-out accuracy over all block inputs. Probe accuracy is in percent. Fine-tuned rows report mean $\pm$ sample standard deviation over seeds.}
\label{tab:sentiment}
\end{table*}

Table~\ref{tab:sentiment-layerwise} reports the sentiment probe results by seed and layer.

\begin{table*}[t]
\centering
\scriptsize
\resizebox{\textwidth}{!}{%
\begin{tabular}{lrrrrrrrrrrrrrr}
\toprule
Condition & Seed & L0 & L1 & L2 & L3 & L4 & L5 & L6 & L7 & L8 & L9 & L10 & L11 & Max \\
\midrule
LM fine-tuning & 13 & 77.5 & 78.9 & 78.0 & 77.3 & 78.8 & 80.9 & 81.4 & 82.4 & 83.9 & 84.5 & 84.6 & 84.2 & 84.6 \\
LM fine-tuning & 42 & 77.5 & 79.1 & 77.9 & 77.5 & 79.2 & 80.9 & 81.5 & 82.7 & 83.5 & 84.4 & 84.2 & 84.3 & 84.4 \\
LM fine-tuning & 73 & 77.5 & 79.1 & 77.4 & 77.5 & 79.3 & 81.0 & 81.5 & 82.5 & 83.8 & 83.9 & 84.0 & 84.0 & 84.0 \\
Mean matching & 13 & 77.6 & 78.1 & 76.8 & 76.5 & 79.0 & 79.7 & 79.8 & 80.8 & 80.4 & 81.3 & 79.7 & 79.4 & 81.3 \\
Mean matching & 42 & 77.8 & 78.0 & 77.4 & 76.7 & 79.1 & 80.0 & 80.0 & 80.3 & 80.8 & 81.3 & 80.0 & 79.8 & 81.3 \\
Mean matching & 73 & 77.5 & 78.5 & 77.5 & 76.2 & 78.8 & 79.6 & 80.3 & 80.8 & 81.0 & 80.6 & 80.0 & 79.8 & 81.0 \\
Covariance matching & 13 & 77.5 & 78.6 & 78.0 & 77.3 & 79.5 & 81.3 & 81.1 & 83.0 & 84.0 & 83.9 & 83.7 & 84.0 & 84.0 \\
Covariance matching & 42 & 77.5 & 78.8 & 78.2 & 77.4 & 79.8 & 81.0 & 81.2 & 82.8 & 83.4 & 83.8 & 83.7 & 83.5 & 83.8 \\
Covariance matching & 73 & 77.5 & 78.8 & 77.8 & 77.0 & 79.3 & 81.2 & 81.0 & 82.8 & 84.0 & 83.7 & 83.7 & 84.2 & 84.2 \\
Mean + covariance & 13 & 77.6 & 78.1 & 77.8 & 76.6 & 79.2 & 79.8 & 80.0 & 81.0 & 80.5 & 81.3 & 79.7 & 79.2 & 81.3 \\
Mean + covariance & 42 & 77.7 & 78.3 & 77.3 & 76.1 & 79.0 & 80.0 & 80.0 & 80.9 & 81.0 & 81.2 & 80.8 & 80.5 & 81.2 \\
Mean + covariance & 73 & 77.5 & 78.0 & 77.5 & 76.7 & 79.0 & 79.6 & 80.2 & 80.7 & 81.0 & 80.8 & 80.0 & 79.1 & 81.0 \\
\bottomrule
\end{tabular}%
}
\caption{IMDb held-out linear probe accuracy by GPT-2 block-input layer and training seed, in percent. Max is the largest held-out layer accuracy within the same seed.}
\label{tab:sentiment-layerwise}
\end{table*}

\section{Transfer and Sensitivity Analyses}
\label{app:sensitivities}

These analyses are intentionally outside the eight-page main argument.
They do not replace the complete Llama causal grid.

\subsection{Frozen-direction cross-dataset transfer}
\label{app:cross-dataset}
This evaluation is fixed before any cross-dataset output is inspected and is excluded from training, direction extraction, model selection, and hyperparameter selection.
The harmful class contains all 400 text behaviors and all 100 separate adversarial-training behaviors released in the official HarmBench repository \citep{mazeika2024harmbench}.
For contextual HarmBench rows, we follow its DirectRequest rendering by prepending the provided context and separator to the behavior.
The harmless class contains all 250 safe XSTest prompts \citep{rottger2024xstest}, plus a seed-42 sample of 25 prompts from each of the ten OR-Bench Hard categories \citep{cui2025orbench}.
The OR-Bench supplement reaches 500 independent harmless examples without duplicating, resampling, or paraphrasing XSTest prompts.

The resulting 1,000-example set is balanced by prompt class and is not behavior-conditioned on any evaluated model.
Construction rejects duplicate exact or whitespace-normalized prompts and asserts zero overlap by standardized sample ID, source ID, exact prompt, and normalized prompt against the WildGuardMix training, development, external, and unfiltered-external splits.
The manifest records the data versions, sampling rules, and results of the overlap checks.

After the matching condition is selected using WildGuardMix development curves only, we evaluate the untouched instruction model, that selected matching checkpoint, and one-epoch ReFAT using their already-frozen WildGuardMix direction sequences at $K\in\{0,1,4,8,16,20\}$.
Generation, WildGuard refusal classification, and 95\% Wilson intervals match the main protocol, and all 500 examples from each class are evaluated.
No cross-dataset output can alter the chosen method, direction sequence, or attack rank.
Table~\ref{tab:cross-dataset} reports cross-dataset transfer.

\begin{table*}[t]
\centering
\scriptsize
\setlength{\tabcolsep}{5pt}
\begin{tabular}{lrll}
\toprule
Model & $K$ & Harmful refusal, \% [95\% CI] & Harmless refusal, \% [95\% CI] \\
\midrule
baseline & 0 & 78.4 [74.6, 81.8] & 35.0 [30.9, 39.3] \\
baseline & 4 & 5.2 [3.6, 7.5] & 2.0 [1.1, 3.6] \\
baseline & 20 & 1.8 [0.9, 3.4] & 4.4 [2.9, 6.6] \\
mean\_kd\_chat & 0 & 73.4 [69.4, 77.1] & 22.8 [19.3, 26.7] \\
mean\_kd\_chat & 4 & 31.2 [27.3, 35.4] & 26.6 [22.9, 30.6] \\
mean\_kd\_chat & 20 & 3.6 [2.3, 5.6] & 2.2 [1.2, 3.9] \\
refat\_1ep & 0 & 98.0 [96.4, 98.9] & 54.0 [49.6, 58.3] \\
refat\_1ep & 4 & 27.4 [23.7, 31.5] & 10.2 [7.8, 13.2] \\
refat\_1ep & 20 & 3.8 [2.4, 5.9] & 1.0 [0.4, 2.3] \\
\bottomrule
\end{tabular}
\caption{Frozen-direction transfer to a balanced 1,000-prompt mixture of HarmBench, XSTest, and OR-Bench. Refusal rates are percentages with 95\% Wilson intervals.}
\label{tab:cross-dataset}
\end{table*}

\paragraph{Unfiltered behavior conditioning.}
We re-evaluate the fixed iterated directions for the untouched model, the selected matching condition, and one-epoch ReFAT at $K\in\{0,1,4,8,16,20\}$ on the balanced but behavior-unfiltered external split in Table~\ref{tab:data-splits}.
Table~\ref{tab:unfiltered-sensitivity} reports refusal rates on the unfiltered split.

\begin{table*}[t]
\centering
\scriptsize
\begin{tabular}{lrrrr}
\toprule
Condition & $K$ & Harm refusal & Harmless refusal & Selectivity gap \\
\midrule
Untouched Instruct & 0 & 90.0 [87.1,92.3] & 13.2 [10.5,16.4] & 76.8 \\
Untouched Instruct & 1 & 32.0 [28.1,36.2] & 4.8 [3.2,7.0] & 27.2 \\
Untouched Instruct & 4 & 12.2 [9.6,15.4] & 2.2 [1.2,3.9] & 10.0 \\
Untouched Instruct & 8 & 9.0 [6.8,11.8] & 2.2 [1.2,3.9] & 6.8 \\
Untouched Instruct & 16 & 4.4 [2.9,6.6] & 1.6 [0.8,3.1] & 2.8 \\
Untouched Instruct & 20 & 3.6 [2.3,5.6] & 2.6 [1.5,4.4] & 1.0 \\
Mean + KD & 0 & 88.4 [85.3,90.9] & 5.8 [4.1,8.2] & 82.6 \\
Mean + KD & 1 & 44.4 [40.1,48.8] & 4.2 [2.8,6.3] & 40.2 \\
Mean + KD & 4 & 38.2 [34.0,42.5] & 23.4 [19.9,27.3] & 14.8 \\
Mean + KD & 8 & 24.8 [21.2,28.8] & 13.8 [11.1,17.1] & 11.0 \\
Mean + KD & 16 & 4.4 [2.9,6.6] & 3.0 [1.8,4.9] & 1.4 \\
Mean + KD & 20 & 3.2 [2.0,5.1] & 2.0 [1.1,3.6] & 1.2 \\
ReFAT (1 epoch) & 0 & 98.0 [96.4,98.9] & 28.6 [24.8,32.7] & 69.4 \\
ReFAT (1 epoch) & 1 & 96.2 [94.1,97.6] & 29.4 [25.6,33.5] & 66.8 \\
ReFAT (1 epoch) & 4 & 39.2 [35.0,43.5] & 4.0 [2.6,6.1] & 35.2 \\
ReFAT (1 epoch) & 8 & 20.8 [17.5,24.6] & 2.0 [1.1,3.6] & 18.8 \\
ReFAT (1 epoch) & 16 & 5.6 [3.9,8.0] & 0.0 [0.0,0.8] & 5.6 \\
ReFAT (1 epoch) & 20 & 6.0 [4.2,8.4] & 0.2 [0.0,1.1] & 5.8 \\
\bottomrule
\end{tabular}
\caption{Sensitivity on the balanced external split without source-model behavior conditioning. Refusal and gaps are percentages; brackets are 95\% Wilson intervals.}
\label{tab:unfiltered-sensitivity}
\end{table*}

\paragraph{Candidate-layer cutoff.}
We rerun extraction for the untouched model and selected matching condition with no source-layer pruning, then evaluate the resulting $K=0,\ldots,20$ curves on the same external split.
This tests whether discarding the final 20\% of candidate source layers materially changes the ranking.
Table~\ref{tab:layer-cutoff-sensitivity} reports results with and without the layer cutoff.

\begin{table*}[t]
\centering
\scriptsize
\begin{tabular}{lllrrrr}
\toprule
Condition & Candidate setting & Source layers & Dev $K_{\mathrm{point}}$ & Ext. $K_{\mathrm{point}}$ & Ext. $K_{\mathrm{conf}}$ & Sel. AUC \\
\midrule
Untouched Instruct & Final 20\% pruned & -- & 2 & 3 & 4 & 0.164 \\
Untouched Instruct & All source layers & 0--15 (16) & 2 & 2 & $>20$ & 0.229 \\
Mean + KD & Final 20\% pruned & -- & 6 & 11 & 11 & 0.247 \\
Mean + KD & All source layers & 0--15 (16) & 2 & 3 & 3 & 0.056 \\
\bottomrule
\end{tabular}
\caption{Candidate-source-layer sensitivity. Both settings freeze development-extracted directions before evaluation on the same external split.}
\label{tab:layer-cutoff-sensitivity}
\end{table*}

\paragraph{Qwen transfer.}
The second-family transfer uses \texttt{Qwen/Qwen2.5-1.5B-Instruct} \citep{qwen2024qwen25}.
Its source prompt pool is \texttt{allenai/wildguardmix}, and response classification uses \texttt{allenai/wildguard}.
The pipeline discards all inherited Llama responses and refusal labels, generates fresh Qwen responses, and classifies the exact Qwen prompt-response pair.
It reuses only the prompt-class label derived from WildGuardMix subcategory and uses the new model-specific refusal label only for behavior conditioning.
We assign prompts to splits by hash before generating responses and record the data versions, split sizes, and overlap checks in the manifest.

The transfer comparison includes untouched Qwen, SFT-only, KD-only, the Llama-selected matching objective, and one-epoch ReFAT under the same response targets and evaluation protocol.
Table~\ref{tab:qwen-transfer} reports the Qwen transfer results.

\begin{table*}[t]
\centering
\scriptsize
\begin{tabular}{lrrrrrr}
\toprule
Condition & Clean harm & Clean harmless & Dev $K_{\mathrm{point}}$ & Ext. $K_{\mathrm{point}}$ & Ext. $K_{\mathrm{conf}}$ & Sel. AUC \\
\midrule
Untouched Instruct & 99.4 [98.3,99.8] & 3.2 [2.0,5.1] & 2 & 2 & 2 & 0.241 \\
SFT only & 97.8 [96.1,98.8] & 5.4 [3.7,7.7] & 1 & 1 & 1 & 0.056 \\
KD only & 97.8 [96.1,98.8] & 7.6 [5.6,10.3] & 2 & 2 & 2 & 0.080 \\
Mean + KD & 97.8 [96.1,98.8] & 10.4 [8.0,13.4] & 1 & 1 & 1 & 0.020 \\
ReFAT (1 epoch) & 99.8 [98.9,100.0] & 38.2 [34.0,42.5] & 4 & 4 & 4 & 0.094 \\
\bottomrule
\end{tabular}
\caption{Qwen2.5-1.5B transfer using fresh model-specific responses and refusal labels. External results use 500 prompts per class; clean rates include 95\% Wilson intervals.}
\label{tab:qwen-transfer}
\end{table*}

\end{document}